# Rethinking deep learning: linear regression remains a key benchmark in predicting terrestrial water storage

**Wanshu Nie[1,2], Sujay V. Kumar[2], Junyu Chen[3], Long Zhao[4], Olya Skulovich[1,2], Jinwoong Yoo[5,2], Justin Pflug[5,2], Shahryar Khalique Ahmad[1,2], Goutam Konapala[6]**

[1] Science Applications International Corporation, McLean, VA 22102

[2] Hydrological Sciences Lab, National Aeronautics and Space Administration (NASA) Goddard Space Flight Center, Greenbelt, MD 20771

[3] Department of Radiology and Radiological Science, Johns Hopkins School of Medicine, MD, 21205

[4] Department of Analytics and Operations, National University of Singapore, Queenstown, Singapore 119245

[5] Earth System Science Interdisciplinary Center, University of Maryland, College Park, MD 20740

[6] Pacific Northwest National Laboratory, Richland, WA 99354

Corresponding author: Wanshu Nie (nwanshu1@jhu.edu)

**Key Points:**

- We compare linear regression, LSTM, and Transformer models for predicting terrestrial water storage at basin scale over the globe.

- Linear regression remains a robust benchmark, outperforming LSTM and Transformer models in various tasks.

- Traditional statistical models and global datasets that capture human and natural impacts are essential for deep learning model evaluation.

**Abstract**

Recent advances in machine learning such as Long Short-Term Memory (LSTM) models and Transformers have been widely adopted in hydrological applications, demonstrating impressive performance amongst deep learning models and outperforming physical models in various tasks. However, their superiority in predicting land surface states such as terrestrial water storage (TWS) that are dominated by many factors such as natural variability and human driven modifications remains unclear. Here, using the open-access, globally representative HydroGlobe dataset – comprising a baseline version derived solely from a land surface model simulation and an advanced version incorporating multi-source remote sensing data assimilation – we show that linear regression is a robust benchmark, outperforming the more complex LSTM and Temporal Fusion Transformer for TWS prediction. Our findings highlight the importance of including traditional statistical models as benchmarks when developing and evaluating deep learning models. Additionally, we emphasize the critical need to establish globally representative benchmark datasets that capture the combined impact of natural variability and human interventions.

**Plain Language Summary**

Recent progress in machine learning has led to the widespread use of deep learning models in studying land freshwater systems, but it remains uncertain if they're always the best tools for such applications. In this study, we use a new, global dataset called HydroGlobe to test different data-driven models. Surprisingly, we find that a basic linear regression model—one of the simplest tools—actually performs better than more complex models like LSTM and Transformers in predicting land water storage. Our results suggest that researchers should always compare deep learning models against simpler traditional statistical benchmarks, and that having high-quality, global datasets that include both natural and human effects is crucial for building better deep learning models.

## 1 Introduction

Terrestrial water storage (TWS) is a key indicator of the world's freshwater availability, encompassing all forms of water stored on and beneath the land surface, including soil moisture, groundwater, surface water, and snow. As a fundamental component of the global hydrological cycle, accurate TWS estimates are essential for applications related to preserving ecosystems, supporting agriculture, and ensuring water and food security for livelihoods. In the past two decades, the ability to measure TWS from satellites has revealed rapid and substantial changes, shaped by the interplay between climate variability and human activities including groundwater extraction, land-use change, and reservoir operations (Getirana et al., 2017; M. Rodell et al., 2018; Matthew Rodell et al., 2024; Sterling et al., 2013). In many regions, these shifts appear to be accelerating at a rate that exceeds the shift driven by anthropogenic climate change, highlighting the increasing influence of human factors on water storage dynamics (Eicker et al., 2016; Humphrey et al., 2016). Improving our ability to model and predict TWS fluctuations is essential for the early detection of hydrological extremes like floods and droughts, sustaining water resources management, and assessing long-term water availability under changing climatic and socio-economic conditions.

With the recent advancements in deep learning (DL), data-driven models are increasingly being explored as alternatives or complements to physical models for hydrology applications (Nearing et al., 2024; Reichstein et al., 2019; Shen et al., 2023; Zhang et al., 2025). Their ability to directly extract meaningful features from observational data, combined with their computational efficiency in processing growing streams of geospatial data, makes them particularly appealing. Given the complex temporal dynamics in hydrological processes, where both long-term patterns (e.g., El Niño-Southern Oscillation) and short-term events (e.g., rapid drying and wetting) from atmospheric and biogeophysical processes influence hydrological predictions, the Long Short-Term Memory (LSTM) networks have become a popular choice in hydrology, such as for rainfall-runoff modeling. Several studies have demonstrated that LSTM consistently outperforms physical models in hydrological applications (Feng et al., 2020; Konapala et al., 2020; Kratzert et al., 2019; Liu et al., 2024; Nearing et al., 2024). LSTM's inherent recurrent structure, computational efficiency, the flexibility in incorporating both time-varying and static covariates, along with its capability to learn from large datasets and generalize information across basins, have made LSTMs particularly effective for sequential hydrological modeling – outperforming traditional process-based models in data-scarce or cross-regional applications. However, LSTMs can be computationally expensive with long input sequences. Their limited ability to capture long-term dependencies when trained on shorter sequences also poses challenges for modeling nonstationary hydrological dynamics.

More recently, Transformer architectures (Vaswani et al., 2017) have attracted considerable attention, driven by their widespread success in natural language processing (Devlin et al., 2019; Lewis et al., 2019). They have also been increasingly applied in hydrological studies, exhibiting mixed performance in hydrological time series prediction (Li et al., 2024; Liu et al., 2024; Rasiya Koya & Roy, 2024; Wei et al., 2023; Yin et al., 2023). Unlike LSTMs, which incorporate an inductive biases tailored to sequential data – favoring patterns that evolve over time – Transformers do not assume such biases and instead relying on self-attention mechanisms to model dependencies across an entire input sequence in parallel. This allows Transformer to learn both long- and short-dependency structures directly from data, without being constrained by predefined temporal structures (Hochreiter & Schmidhuber, 1997; Vaswani et al., 2017). While this flexibility of Transformers may enhance predictive capability in complex, dynamically evolving systems, recent studies have raised questions on whether Transformers are truly effective for time series tasks (Elsayed et al., 2021; Tan et al., 2024; Zeng et al., 2023). In particular, Transformer's self-attention mechanism is inherently permutation-invariant, meaning it treats inputs the same regardless of their order (Vaswani et al., 2017). While positional encodings and time embeddings are used to mitigate this, there are ongoing discussions and research suggesting these additions may still lead to a loss of temporal information, potentially undermining its effectiveness in time-dependent contexts compared to sequence-aware architectures (Wu et al., 2021; Zeng et al., 2023; Zhou et al., 2021). More broadly, these concerns reflect a growing debate on whether DL models are always necessary or optimal for time series analysis, as some studies suggest that simpler models with careful design choices can achieve comparable or even superior performance compared to DL approaches (Bergmeir et al., 2023; Chuang et al., 2024; Elsayed et al., 2021; Grinsztajn et al., 2022).

In this paper, we systematically assess whether advanced DL approaches offer significant advantages in hydrological time series prediction, using TWS as the target variable. To do this, we employ a simple linear regression model as our benchmark and investigate whether widely used LSTM and the more recent Temporal Fusion Transformer (TFT; Lim et al., 2019) – can improve in predicting TWS. Notably, TFT combines the strengths of self-attention mechanisms with local temporal processing provided by integrated LSTM blocks and is designed to flexibly incorporate both time-varying and categorical covariates representing a significant advancement in the field. Previous studies have shown that TFT is not only competitive within deep learning models (Huy et al., 2022; B. Wu et al., 2022), but has also demonstrated strong performance against simpler yet well-designed traditional machine learning approaches, such as gradient boosting trees (Elsayed et al., 2021). In a task of streamflow prediction, TFT has been reported as the best-performing model compared to state-of-the-art LSTM and vanilla Transformer when applied to the Caravan – a global hydrological benchmark dataset (Rasiya Koya & Roy, 2024). In this study, we train our linear regression, LSTM, and TFT models on HydroGlobe (https://ldas.gsfc.nasa.gov/hydroglobe)(Nie et al., 2024), an open-access, globally representative land water reanalysis dataset, to assess and compare the performance of the models.

Beyond its baseline version (hereinafter referred to as the open loop (OL) dataset) that relies solely on land surface model simulation, the advanced version (hereinafter referred to as the DA dataset) for HydroGlobe integrates satellite-based observations of TWS, soil moisture, and leaf area index with model simulations through data assimilation within NASA's Land Information System (LIS) modeling framework (Kumar et al., 2006), covering the period of 2003-2020 at a spatial resolution of 10 km. This integration offers a significant advantage over purely observation-based datasets in that it provides global, spatially and temporally continuous estimates of critical hydrological states and fluxes. Furthermore, the representation of real-world land surface conditions is improved by the integration of information from satellite observations from multiple sources, particularly in capturing hydrological responses from human land and water management.

In our previous work (Nie et al., 2024), we demonstrated the simulation of evapotranspiration, groundwater, and gross primary production is significantly enhanced by HydroGlobe with data assimilation when compared to the baseline HydroGlobe dataset. Notably, we found that TWS in the DA dataset exhibits stronger nonstationarity, characterized by the presence of long-term trends and shifts in seasonality, as well as changes in the frequency of extreme events, largely attributable to human intervention in water resources. Nonstationary environments, where the probabilistic properties of data change over time, pose great challenges for data-driven models trained under stationary assumptions. If such models fail to account for shifting distributions, their predictive performance may degrade over time, leading to suboptimal results at best or catastrophic failure at worst (Ditzler et al., 2015). A common practice in hydrological time series prediction is to split datasets into training and testing sets based on the time period. However, when time series exhibit strong nonstationarity, this practice may introduce challenges related to covariate shifts, transfer learning, and domain adaptation – each of which involves shifts from training to testing probability distributions (Ditzler et al., 2015). In this study, beyond investigating how deep learning models compare to the simple linear benchmark in predicting TWS, we also evaluate model performance on both the OL and DA datasets. While DA datasets reflect more realistic system states as it incorporates information from satellites that are capable of detecting human interventions on land surface states, the contrast between OL and DA provides a valuable opportunity to assess how different models learn and generalize under varying degrees of nonstationarity. This comparison is included throughout our experiments to better understand model robustness and adaptability to increasingly complex and dynamic conditions.

Given the rapid evolution of DL methods and the increasing enthusiasm to adapt them for applications in hydrology, we consider it essential to assess the effectiveness of these methods against simple, yet well-established approaches to ensure meaningful progress in the field. Furthermore, we advocate for the development and utilization of globally representative datasets such as HydroGlobe, which comprehensively capture both climatic and anthropogenic influences, yet are currently underrepresented in global benchmark datasets for developing and assessing DL models. Developing models that accurately simulate hydrological shifts from both natural and human-driven changes is a key to crafting more informed policies and strategies that support sustainable economic well-being amidst rapid global change.

## 2 Materials and Methods

### 2.1 Datasets

This study utilizes input features and target variables derived from NASA's HydroGlobe dataset (Nie et al., 2024). HydroGlobe is a land surface reanalysis that integrates several remote-sensing datasets with the Noah-MP land surface model (Niu et al., 2011) using the NASA LIS software framework (Kumar et al., 2006). We employ two distinct versions of HydroGlobe simulations:

1. OpenLoop (OL) - A baseline simulation based solely on the Noah-MP model, without observational constraints from data assimilation.

2. Data Assimilation (DA) - A reanalysis dataset incorporating multi-source remote-sensing based observations assimilated into the Noah-MP model, including GRACE mascon-based terrestrial water storage (TWS) anomalies (Loomis et al., 2019), the European Space Agency's (ESA CCI) active and passive combined surface soil moisture (SSMC) (Dorigo et al., 2017), and the MODIS-based leaf area index (LAI) (Myneni et al., 2015).

As described in Nie et al.(2024), the OL dataset primarily captures hydrological response to natural variabilities, whereas the DA dataset, enhanced by the remotely sensed information, better represents the combined impact of climate variability and anthropogenic activities on the terrestrial water cycle. Consequently, TWS in the DA dataset exhibits stronger nonstationary patterns, including long-term trends, shifts in seasonality, and changes in extremes, compared to the OL dataset.

These input and target variables below are used in our data-driven models in this study as in the Models section following:

- Meteorological time-varying inputs: Precipitation from NASA's Integrated Multi-satellitE Retrievals for Global Precipitation Measurement (IMERG; Huffman et al., 2015) final run version V06B; 2-m air temperature from NASA's Modern-Era Retrospective Analysis for Research and Applications, version 2 (MERRA-2; Gelaro et al., 2017). These two variables are also used as meteorological drivers for HydroGlobe simulations.

- Land surface time-varying input: LAI and SSMC simulated by HydroGlobe (both OL and DA simulations).

- Static inputs: Elevation and slope from the Multi-Error-Removed Improved-Terrain (MERIT; Yamazaki et al., 2017), sand, silt, and clay fraction from the International Soil Reference Information Centre (ISRIC; Hengl et al., 2017), forest and cropland fraction from the Moderate Resolution Imaging Spectroradiometer – International Geosphere Biosphere Program (MODIS-IGBP) land cover dataset (Friedl et al., 2010), which are also static inputs for both OL and DA simulations. We also included basin area and monthly climatology of precipitation, temperature, and LAI as static inputs, totaling up to 11 features.

- Target variable: TWS obtained from HydroGlobe (both OL and DA simulations).

Note that SSMC, LAI, and TWS differ remarkably between the OL and DA simulations due to the impact of data assimilation. In OL simulation, LAI primarily relies on the prognostic vegetation module in Noah-MP physics, whereas LAI in DA is directly influenced by the assimilation of MODIS LAI observations and indirectly affected by the assimilation of soil moisture and TWS through vegetation-water interaction. Similarly, SSMC and TWS in DA are directly affected by the assimilation of GRACE TWS anomalies and ESA CCI surface soil moisture, while also being indirectly affected by the assimilation of LAI.

The original HydroGlobe dataset is available at a 10 km spatial resolution with a daily time scale spanning 2003-2020. In this study, we aggregate the data into monthly and daily basin-averaged time series using WMO's HydroSHED (Lehner & Grill, 2013) basin polygons, covering 515 basins over the globe. Importantly, we did not use TWS data as input features in our data-driven models for two reasons: 1) near real-time TWS for DA is unavailable due to the ~3-month latency in the GRACE product, and 2) our objective is to evaluate the model's predictive skill without relying on lagged TWS values (i.e., autoregressive information) as inputs.

## 2.2 Models

Benchmarks play a crucial role in evaluating the performance of a proposed model for a specific application. While recent studies often compare DL methods exclusively against other DL models (Ghobadi & Kang, 2022; Liu et al., 2024; Koya & Roy, 2024; Yin et al., 2023), we emphasize the importance of benchmarking against well-established traditional machine learning methods for hydrological applications. Such comparisons not only ensure robustness and practical relevance but also provide clearer insights into the actual improvement that these methods provide over simpler and more explainable modeling approaches. Given the recent surge in the development of DL-based sequence modeling architectures, including various LSTMs and Transformer models, it is essential to include simple, yet informative benchmarks rather than limiting evaluation to a pool of deep learning models alone (Hewamalage et al., 2023). This ensures a more comprehensive and meaningful assessment.

For our baseline, we employ linear regression to predict TWS. We consider two versions of the linear model:

Linear_single (benchmark): A linear regression model built separately for each basin. This model serves as a benchmark to evaluate all other proposed models.

Linear_glob: A linear regression model built using data from all basins. Given the heterogeneity among basins, the hypothesis is that aggregating data from diverse distributions into a linear model may lead to poorer performance compared to training linear models separately for each basin.

In both cases, only time-varying features are used for predicting TWS. The data is split into training (2003-2015) and testing (2016-2020) periods, and each basin's data is standardized individually using a standard scaler computed from the training data. The linear models incorporate three types of features: 1) lagged values for precipitation, temperature, LAI, and SSMC corresponding to the sequence length; 2) monthly seasonal categorical variables (with one month omitted to avoid collinearity); and 3) a trend feature defined by the time index – an index of the position of values in the full time series. The monthly categorical variables are encoded as one-hot indicators that represent different months to serve as a proxy for recurring seasonal effects. For example, with a sequence length of 12 months, the feature vector comprises 48 lagged time-varying features, 11 monthly categorical variables, and 1 trend feature, totaling 60 dimensions.

For DL experiments, we select two state-of-the-art models widely used by practitioners:

Long Short-Term Memory (LSTM) network: We adopted a single layer LSTM model that processes both time-varying and static inputs. This model has been widely used in the hydrology field for tasks such as rainfall-runoff modeling and streamflow prediction, where it has been shown to outperform several popular hydrological models or physical model-based early warning systems (Feng et al., 2020; Konapala et al., 2020; Kratzert et al., 2019; Nearing et al., 2024).

Temporal Fusion Transformer (TFT): TFT leverages a hybrid architecture that integrates LSTM units with multi-head attention mechanisms. We choose TFT as a representative of transformer-based models, as 1) this design enables the model to capture both long-term dependencies and fine-grained temporal dynamics, addressing the limitations associated with pure attention-based approaches that can be permutation invariant; 2) TFT accepts both static and time-varying covariates, whereas many popular transformers are only developed for univariate or multivariate time series tasks (Wang et al., 2024).

DL models benefit from "data synergy", where larger and more diverse datasets can lead to a more robust model (Fang et al., 2022; Kratzert et al., 2022). To leverage this effect, both LSTM and TFT are trained on global data from all basins rather than on individual basin data to improve generalization and mitigate overfitting. This approach aligns with recent recommendations against training LSTMs on individual basins(Kratzert et al., 2022) and supports findings that highlight the scaling properties of Transformers, where larger datasets lead to improved performance across tasks (Kaplan et al., 2020; Zhai et al., 2022). The data is split into training (2003-2012), validation (2013-2015), and testing (2016-2020) periods. Both static and time-varying features are standardized for each basin separately using the training period statistics before being merged into a global training set. Hyperparameters for LSTM and TFT models are optimized using Optuna (Akiba et al., 2019) over 50 trials (with each trial running for 10 epochs) on the task of predicting one monthly step of TWS using a 12-month sequence. Table 1 provides the details for the final hyperparameter configurations. To prevent overfitting, we implement an early stopping strategy that monitors the validation loss. Specifically, if the validation loss does not improve by at least 0.0001 over 10 consecutive epochs, training is halted, and the model at the epoch when the minimum validation loss is reached is saved and used for evaluation. The same set of parameters is applied for additional experiments that require LSTM or TFT, such as those with different sequence lengths or forecast steps. We note that this may suffer from shortcomings, as the hyperparameter set optimal for one experiment might not be ideal for another. Nevertheless, given that the core data dynamics are preserved, and the underlying features and target remain consistent, any performance differences are expected to be minimal relative to the benefits of computational efficiency.

**Table 1.** Hyperparameters for LSTM and TFT models.

| Models | LSTM | | TFT | |
|---|---|---|---|---|
| Datasets | OL | DA | OL | DA |
| Parameters and Description | | | | |
| **num_layers:** Number of stacked LSTM layers | 1 | 1 | 1 | 1 |
| **hidden_size (optimized):** Number of neurons in the LSTM layer for LSTM or that in variable selection, LSTM, GRN, and attention blocks for TFT | 512 | 64 | 80 | 80 |
| **nheads (optimized):** Number of attention heads in the multi-head attention block | n/a | n/a | 5 | 4 |
| **Intial learning rate (optimized):** Initial magnitude of parameter updates during training | 0.0016 | 0.0097 | 0.0016 | 0.0011 |
| **dropout (optimized):** Fraction of deactivated random neurons during training | 0.2 | 0.1 | 0.2 | 0.2 |
| **weight initialization method (optimized):** | xavier | orthogonal | n/a | n/a |
| **Early stopping val_loss min_delta:** minimum change in the validation loss required to register an improvement | 0.0001 | 0.0001 | 0.0001 | 0.0001 |
| **Early stopping patience level:** the number of consecutive epochs without such improvement that triggers early stopping | 10 | 10 | 10 | 10 |

While our primary objective is to assess whether DL models are essential for hydrological TWS prediction and to compare their performance against a traditional linear benchmark, it is important to recognize that linear models may not fully capture the potential complex, nonlinear relationships between features and the target, if there are any. Consequently, we extend our investigation by incorporating two state-of-the-art tree-based models that are well-suited to modeling nonlinear dynamics:

Random Forest (RF; Breiman, 2001): an ensemble method that aggregates predictions from multiple decision trees to reduce overfitting and enhance robustness.

Light Gradient-Boosting Machine (LightGBM; Ke et al., 2017): a gradient boosting framework that constructs trees sequentially and with optimized learning. Compared to the traditional gradient boosting method, which is applied level wise, LightGBM reduces training time and memory usage by employing a histogram-based algorithm and a leaf-wise tree growth strategy.

Both RF and LightGBM are applied at the basin level and are optimized separately for each basin via grid search on key parameters. The features and target, as well as the train-test split are identical to those used for Linear_single. Table 2 summarizes the list of parameters and their search ranges.

**Table 2.** Parameter and its search range for optimization in RF and LightGBM models for each basin.

| Models | Parameters | Description | Search Range |
|---|---|---|---|
| RF | n_estimators | Number of trees in the forest | 10, 50, 100 |
| | max_depth | Maximum depth of each tree | 5, 10, None |
| | min_samples_split | Minimum number of samples required to be at a leaf node | 2, 5, 10 |
| | min_samples_leaf | Minimum number of samples required to be at a leaf node | 2, 5, 10 |
| LightGBM | n_estimators | Number of boosting iterations | 10, 50, 100 |
| | max_depth | Maximum depth of each tree | 3, 5, 7 |
| | learning_rate | Boosting learning rate | 0.01, 0.05, 0.1 |
| | num_leaves | Maximum number of leaves per tree | 10, 20, 30 |
| | min_child_samples | Minimum number of samples required in a child node | 20, 30 |
| | min_gain_to_split | Minimum loss reduction required to perform a split | 0.01, 0.05 |

To ensure a fair comparison across all models, we employ a quantile-based loss function (as TFT only supports quantile loss) and we report our primary results based on quantile loss at the median, equivalent to the mean absolute error loss. Model performance is assessed using a suite of metrics, including bias, RMSE, correlation, and two widely used metrics in the hydrology field: Nash-Sutcliffe efficiency (NSE; Nash & Sutcliffe, 1970) and Kling-Gupta efficiency (KGE; Gupta & Kling, 2011). Below we provide the equation for NSE and KGE calculation. For both metrics, values closer to 1 indicate better performance. Statistical significance is evaluated using a two-sided Mann-Whitney U test at a 5% significance level, and all box plots show distribution quartiles with error bars that represent the full range of the data, excluding outliers.

$$NSE = 1 - \frac{\sum_{t=1}^{T}(y_{pred} - y_{true})^2}{\sum_{t=1}^{T}(y_{true} - \overline{y_{true}})^2} \quad (1)$$

$$KGE = 1 - \sqrt{(r-1)^2 + (\frac{\sigma_{pred}}{\sigma_{true}} - 1)^2 + (\frac{\mu_{pred}}{\mu_{true}} - 1)^2} \qquad (2)$$

In which $y_{pred}$ and $y_{true}$ are predicted TWS and HydroGlobe-based TWS respectively, $\overline{y_{true}}$ is the algorithm mean of $y_{true}$ for the test period. T is the length of the test period, $r$ is the pearson correlation between $y_{pred}$ and $y_{true}$, $\frac{\sigma_{pred}}{\sigma_{true}}$ and $\frac{\mu_{pred}}{\mu_{true}}$ are the ratio of standard deviation and mean for $y_{pred}$ and $y_{true}$ during the test period.

**2.3 Experiments**

We primarily assess model performance using a regression task, where monthly input features spanning the past 12 months are used to predict the current month's target TWS. To further analyze model behavior under different temporal settings and prediction challenges, we conduct additional experiments, including:

- Regression tasks with sequence lengths ranging from 6 to 18 months with monthly inputs and target.

- Forecasting task at forecast steps ranging from 1 to 6 months ahead using a fixed 12-month sequence length with monthly inputs and target.

- Regression tasks using daily input with a sequence length of 365 days. These experiments are performed to explore the feasibility of using higher temporal resolution data. Since predicting a single-day TWS value may introduce excessive noise, we applied a 30-day moving average kernel to target TWS, where the predicted TWS represents the average over the last 30 days of the sequence window.

**3 Results**

**3.1 Primary results**

The primary results are based on regression tasks applied separately to the global OL and DA dataset aggregated at the monthly, basin-averaged level. To evaluate model performance, we use a linear regression model built for each individual basin (Linear_single) as the benchmark. This is compared against three models trained on global data: a global linear regression model (Linear_glob), an LSTM, and the TFT. Figure 1 presents the cumulative distribution function (CDF) of each model's performance across various evaluation metrics, assessed on test period (2016-2020) for basin-average TWS across 515 basins globally.

For the OL dataset, the benchmark model Linear_single demonstrates significantly better performance across all evaluation metrics, except for bias. The performance of the models is comparable for the bias metric, as Linear_single does not exhibit a significantly smaller bias than the other three models.. In fact, the bias from TFT is significantly smaller than that of Linear_single (based on the one-sided test). Among the competing models, TFT appears to have the most comparable performance to Linear_single overall, followed by LSTM.

Linear_glob ranks the lowest, particularly for the correlation and NSE statistics, highlighting that while a linear model may sufficiently describe basin-specific relationship between the model input features and TWS, fitting a linear model on all basins globally is inadequate due to the substantial heterogeneity in climatological and hydrological characteristics across basins.

For the DA dataset, Linear_single again outperforms the other three models, suggesting that it better captures the nonstationary behavior of TWS, despite relying primarily on the specification of a linear trend and monthly dummies. While TFT benefits from incorporating a global time embedding and captures some degree of nonstationarity (see examples in additional experiments), it does not improve on Linear_single in overall performance, and Linear_glob remains the lowest-performing model. Figure 2 presents the spatial distribution of the best and second-best models amongst Linear_single, LSTM, and TFT, and shows the distribution of NSE metric, stratified by the full-period linear TWS trends (see inset scatter plots). It is evident that the TWS trends are considerably more negative in the DA dataset compared to OL, reflecting a form of stronger nonstationarity. The best performing models for basins with such trends are predominantly Linear_single or TFT. In contrast, LSTM struggles to predict trends in basins exhibiting such strong nonstationarity.

Interestingly, despite Linear_single achieving the best performance across both datasets, its skill in describing the feature-target relationship is notably weaker for the DA dataset as compared to the OL dataset, a pattern observed across all models tested. This discrepancy may be attributed to either missing features related to human impacts on TWS variations or the presence of complex and nonlinear relationships between existing features and the target variable that cannot be adequately captured by simple linear models. For instance, groundwater pumping for irrigation is the main reason for TWS depletion in many managed regions such as northwestern India and southern High Plains in U.S. While such human activity is often climate-driven, particularly in areas where irrigation supplements rainfed agriculture (Asoka et al., 2017; Nie et al., 2021; Russo & Lall, 2017), the relationship between climate conditions, irrigation water withdrawal, and vegetation growth may not be monotonic. Extreme dry conditions, for example, can lead to the termination of cropping and irrigation, deviation from a simple "less rain, more irrigation" relationship. In these scenarios, although climate factors (precipitation, temperature), vegetation indices (LAI), and soil moisture conditions (SSMC) still carry information in implicating TWS change through their influences on groundwater use, this complex interaction is not readily captured by a simple linear model. Given the limited availability of explicit features representing human interventions (e.g., time-varying irrigation withdrawals), DL models – which can better uncover complex nonlinear dynamics directly from the data – may offer greater advantages as compared to linear models. However, their development depends on high-quality datasets that better reflect real-world conditions, such as DA datasets that integrate process-based modeling with satellite observations.

Despite the overall suboptimal performance of DL-based models, TFT most closely matches the performance of Linear_single, making it a potential candidate for a foundation model (Bommasani et al., 2021; Liang et al., 2024) in hydrological applications. A globally pretrained TFT model could potentially be fine-tuned for specific basins or adapted to challenges such as improving prediction for data-scarce or high-uncertainty regions. This approach could reduce the amount of fine-tuning data required for transfer learning while enhancing performance. These findings underscore the potential for future research into developing scalable and efficient DL models for hydrological modeling through strategies like pretraining and transfer learning.

Overall, the approximation error for Linear_single on the OL dataset is sufficiently small that the potential gains from using more complex DL models are limited. The DA dataset presents greater approximation error, likely due to missing features, increased nonlinearity, and concept drift. However, the expected reduction in estimation error from using more sophisticated models remains limited, as the fundamental challenges in the DA dataset outweigh the potential benefits of increased model complexity. Consequently, Linear_single continues to outperform DL models in this task. These findings align with prior studies indicating that advanced DL-based models do not consistently surpass simple linear models (Zeng et al., 2023). While differences in training strategies – individual basins vs. global – must be considered, Linear_single, despite its simplicity, remains a robust benchmark and yields the best performance. This raises the broader question: can DL models provide added utility compared to well-established traditional statistical approaches? Given the inherent linearity in the relationship between the input features and TWS at aggregated basin level, complex DL models may not offer substantial advantages. In the following sections, we further investigate whether DL models can outperform this benchmark under different conditions designed to potentially highlight their strengths, such as varying sequence window lengths, increasing temporal granularity, and transitions from regression to forecasting tasks.

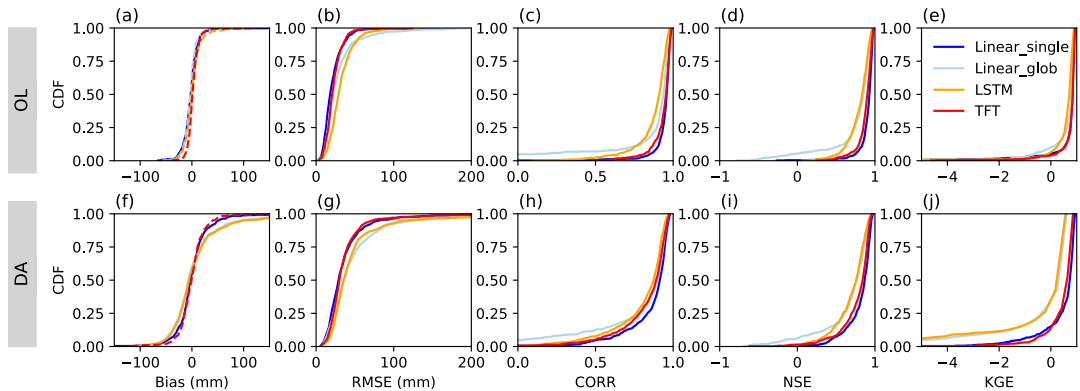

**Figure 1**. Skill of the four data-driven models evaluated across five metrics for the OL (top row) and the DA (bottom row) dataset. Using Linear_single as a benchmark, solid lines indicate that a model performs significantly worse than Linear_single while dashed lines indicate not significantly worse (Mann-Whitney one-side U test, 95% significance level). Additionally, we conducted the other side test for models not significantly worse than Linear_single, finding that only TFT outperforms Linear_single in terms of bias for OL dataset.

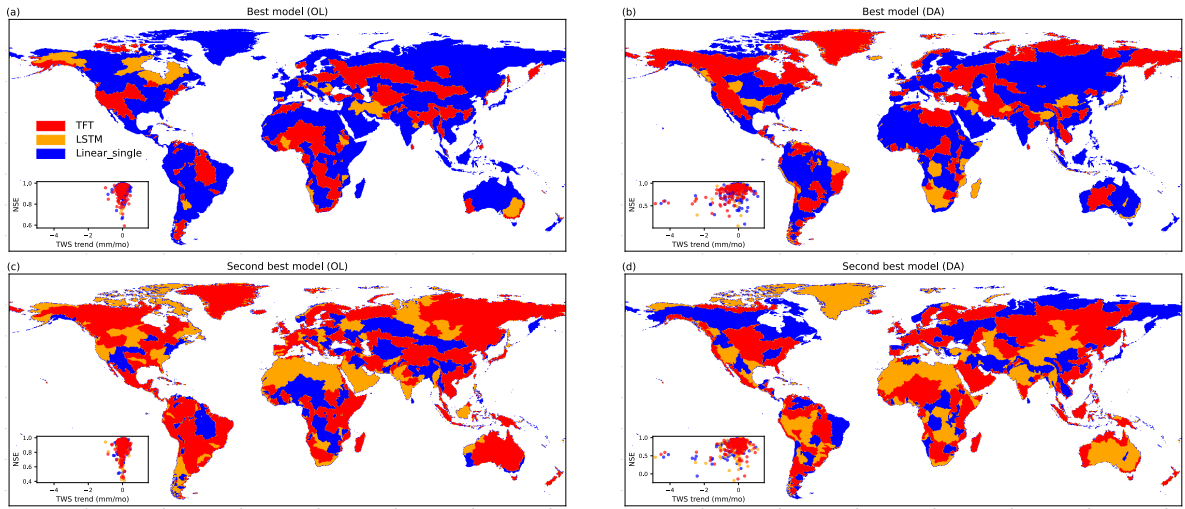

**Figure 2**. Spatial distribution of (**a, b**) the best and (**c, d**) second-best model across all basins for the OL and DA dataset, ranked amongst Linear_single, LSTM, and TFT based on the NSE metric. The inset scatter plots illustrate the relationship between the TWS trends (linear trend over the full period) and NSE metric for the corresponding model choice.

### 3.2 Why does Linear_single perform much better than Linear_glob?

In the context of linear modeling, a larger dataset typically reduces variance while introducing minimal additional bias, provided that the underlying feature-target relationships are consistent across samples. However, the benefit of global aggregation is limited when data points exhibit high heterogeneity, as this assumption breaks down. In our study, a total of 60 features are used for both Linear_single and Linear_glob. These included present and lagged information on precipitation, temperature, LAI, and SSMC represented by 48 features,11 features are monthly categorical variables, and the last one is a trend feature. Within this linear framework, a key question under a linear structure is, whether a globally trained model can effectively generalize across diverse basins, and if not, why does it fail?

To assess this, we compare the distribution of learned coefficients from Linear_single for all basins with those from Linear_glob. As shown in Figure 3, the learned coefficients from Linear_single exhibit a widespread distribution across basins, suggesting that feature-target relationships vary significantly. This high degree of heterogeneity in the coefficient distribution in basin-specific relationships implies that a single set of coefficients learned by Linear_glob introduces the presence of a large bias for individual basins, leading to poor generalization. Consequently, from a bias-variance perspective, pooling data across these heterogeneous basins fails to reduce variance meaningfully to yield performance improvements because the substantial bias introduced by the restrictive global linear structure outweights the potential reduction in variance from increased data volume.

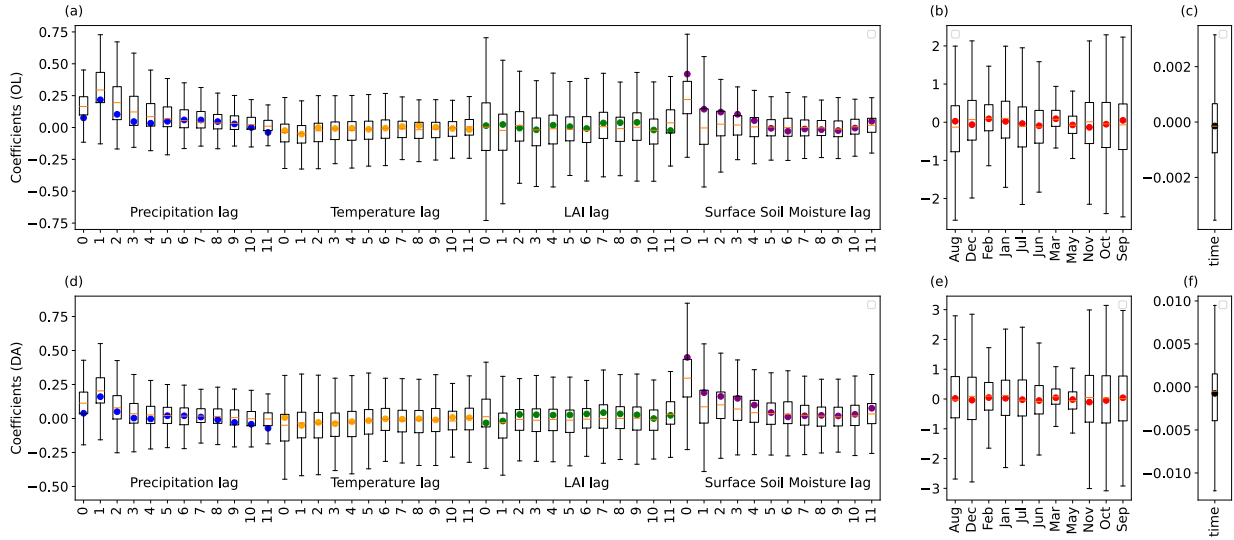

**Figure 3**. Coefficients distribution for all features in Linear_single for all basins (boxplots) vs. Linear_glob (dots) for (**a, b, c**) OL and (**d, e, f**) DA.

### 3.3 Would LSTM and TFT benefit from longer input sequences?

The sequence length plays a critical role in regression accuracy, as it determines the extent of historical patterns that the model can leverage. This consideration is particularly relevant for TWS, where different water storage components exhibit varying degrees of memory of past conditions. For instance, surface soil moisture, as part of TWS, has a relatively short memory due to rapid evaporation and infiltration, whereas groundwater and snowpack dynamics are influenced by climate conditions from months earlier. Moreover, longer sequences capture more information on nonstationarity, preserving information on potential shifts in the mean and variance of the data over time. Theoretically, a sufficiently expressive time series model with strong temporal feature extraction capabilities should benefit from extended sequence lengths (Zeng et al., 2023).

To investigate the impact of input sequence length, we conduct experiments with input sequence lengths of L $\epsilon$ {6, 9, 12, 15, 18} months. The maximum sequence length is capped at 18 months to mitigate overfitting risk in Linear_single, given the limitations in training data. The results indicate that, contrary to expectation, both LSTM and TFT, despite their recurrent and/or attention mechanism, do not exhibit significant performance improvements with increasing sequence length. Although a few basins (less than 5%) exhibit a significant increase in correlation or NSE with longer sequences for both LSTM and TFT across both OL and DA datasets (not shown), the overall distribution of all metrics across all basins does not show a significant shift with increased sequence length (Mann-Whitney U test, Figure 4).

To further understand the contribution of different input time steps to LSTM's predictions, we used the OL dataset as an example and employed the SHAP explainer(Roth, 1988) to compute averaged feature importance across sequence steps (Figure 5a). The results show that, on average, more recent time steps contribute the most to predictions, particularly the closest time step (sequence step = 18), while earlier time steps contribute less, even in longer sequences. However, we observe a slight increase in SHAP values for earlier time steps when the sequence length extends to 15 or 18 months, suggesting that LSTM may begin to leverage information from more distant past steps in longer sequences. Nonetheless, the contribution from these distant steps remains considerably lower than that of the most recent steps. We also analyze the attention weights for TFT (Figure 5b) to understand how the model distributes importance across time steps. The results indicate that TFT's attention patterns vary substantially with sequence length. For sequence lengths of 6 and 9 months, attention is drawn more strongly to the earliest and latest months. With a sequence length of 12 months, contributions decrease as time steps become more distant. However, for longer sequences (i.e., 15 and 18 months), attention weights tend to distribute more towards distant time steps rather than concentrating on the most recent past. This observed inconsistency in attention distribution across varying sequence lengths likely explains why TFT does not show improved performance with longer sequence input. The only robust pattern we observe is for Linear_single: its performance significantly degraded when sequence length increases from 9 (for OL) and 12 (for DA) months to 18 months. This degradation may result from potential overfitting, reflecting a trade-off between available data points and sequence length. Despite these differences, neither LSTM nor TFT surpasses the Linear_single benchmark across both OL and DA datasets for most tested sequence length choices, suggesting that the advantage of complex architectures over linear models remains dataset-dependent.

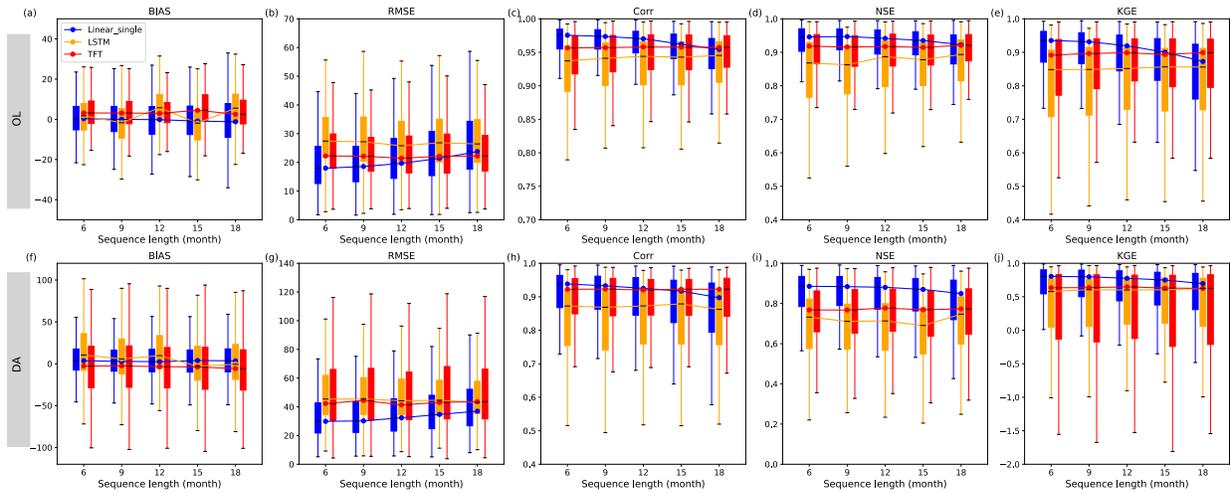

**Figure 4**. Evaluation metrics of Linear_single, LSTM, and TFT with different sequence lengths on OL (top row) and DA (bottom row) dataset.

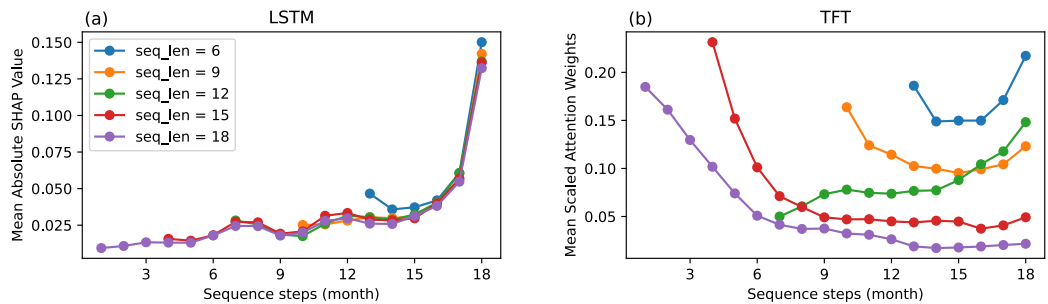

**Figure 5**. (**a**) Mean of SHAP values for OL LSTM, (**b**) Mean scaled attention weights for OL TFT over all time-varying features and across all global basins for models trained on different sequence lengths.

### 3.4 How does LSTM and TFT compare to linear benchmarks in forecasting tasks?

Beyond the regression task, we also investigated the forecasting performance of LSTM and TFT compared to Linear_single. These experiments used a fixed input sequence length of 12 months and vary the forecasting lead time from 1 to 6 months. Unlike the previous regression setting, which followed a sequence-to-one format (predicting a single target month), the forecasting task adopts a sequence-to-sequence format, where multiple future time steps are predicted simultaneously. As expected, the performance of all models deteriorates as the forecasting lead time increases (Figure 6), suggesting that these models primarily capture near-term temporal dependencies rather than long-range patterns, at least when not provided with auto-regressive information on the target variable or known future features, such as forecasted precipitation and temperature. Interestingly, Linear_single remains the most skillful model for short-term forecasting (up to three months), outperforming both DL models. TFT exhibits the most stable performance across different forecasting steps or lead times, surpassing Linear_single when forecasting beyond 3 months. LSTM performs the worst, both in terms of average predictive accuracy and consistency across basins, indicating its limitation in handling longer-term dependencies effectively in this setting. The faster decline in the performance of linear models for longer forecasting leads can be attributed to their inherent limitations in capturing complex temporal dependencies. Linear_single relies on fixed-weight relationships between past inputs and future outputs, which may hold over short leads but fail to account for delayed, nonlinear interactions present in hydrological systems. In contrast, TFT, with its ability to leverage memory and attention mechanisms to dynamically weight past observations based on their relevance to the forecasting target is better equipped to capture these more complex and time-varying relationships, leading to more stable performance at longer forecast leads.

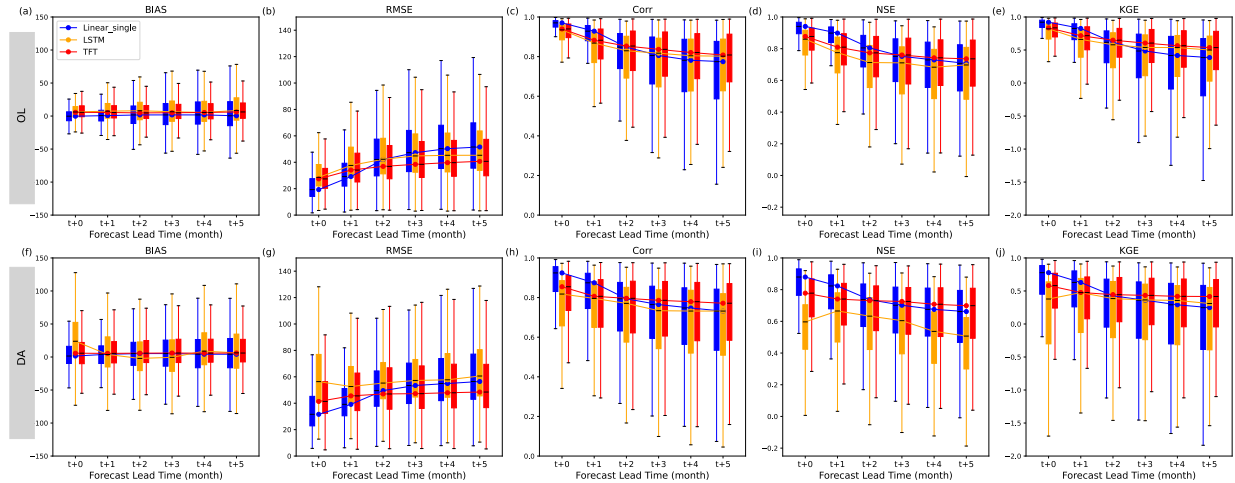

**Figure 6**. Evaluation metrics of Linear_single, LSTM, and TFT with different forecasting lead time on OL (top row) and DA (bottom row) dataset.

### 3.5 What makes TFT learn better in a nonstationary environment?

Understanding how DL models handle nonstationary environments is critical for assessing their suitability in hydrological applications. The DA dataset, which incorporates multi-source satellite measurements, better captures both climate influences and human interventions, resulting in more complex nonstationary behavior compared to the OL dataset. Figure 1 suggests that while TFT achieves slightly lower RMSE and higher KGE than Linear_single, the differences are statistically insignificant. But both models perform much better than LSTM and Linear_glob. One potential explanation for Linear_single's strong performance lies in its explicit linear trend feature, which, although simplistic, provides a direct means of modeling long-term tendencies. In contrast, TFT does not incorporate an explicit trend term but instead relies on learned representations to model temporal dynamics. To understand which components enable TFT to learn in a nonstationary environment, we conduct a data denial study by removing the time_idx feature – a global timestamp embedding that provides a reference point for each sequence within the full time series.

As shown in Table 3, for the group of basins that exhibit significant depletion trends, removing this temporal embedding leads to a substantial degradation in TFT's performance, particularly in bias, RMSE, and KGE. This suggests that the primary mechanism allowing TFT to capture nonstationarity is not its self-attention mechanism but rather the explicit encoding of time via global embeddings. Figure 7 further illustrates this effect by showcasing time series for sample basins with relatively stable TWS (e.g., the Congo basin) against those with clear depletion trends (e.g., the Central basin and Arabian Peninsula). This reinforces that, in the absence of this temporal embedding, TFT struggles to learn and apply nonstationary information effectively. Because self-attention is inherently permutation-invariant, it does not inherently encode temporal ordering (Zeng et al., 2023), leading to information loss in datasets with strong nonstationary trends. Although incorporating global timestep embeddings enhances TFT's ability to represent long-term variations, it does not surpass the effectiveness of a simple linear trend assumption in Linear_single. Moreover, despite the inclusion of temporal embeddings, TFT still underperforms Linear_single across most evaluation metrics, highlighting its limitation in capturing fine-grained temporal relationships.

It is important to note that the better performance of Linear_single should not be interpreted as evidence that a simple linear trend is sufficient to handle nonstationarity in all cases. This assumption may hold when nonstationary behavior is dominated by relatively smooth monotonic trends, such as basins where TWS is under relatively steady depletion due to groundwater pumping for irrigation. However, this assumption is likely to break down under more complex, nonlinear, or abrupt shifts in system dynamics. Effectively learning in nonstationary environments requires careful consideration of model architecture, informed feature selection, and appropriate data pre-processing strategies. Future work could explore how different architecture choices, expanded temporal features, or adaptive learning frameworks can enhance model robustness in dynamically evolving hydrological systems.

**Table 3**. Mean of the metrics for results from TFT trained with and without the global time index embedding for basins with DA dataset that have significant negative trends (201 out of 515 basins).

| Metrics | TFT | TFT_no_timeidx |
|---|---|---|
| \|Bias\| | 24.91 | 41.41 |
| RMSE | 49.05 | 60.11 |
| Corr | 0.79 | 0.79 |
| NSE | 0.74 | 0.72 |
| KGE | 0.29 | -0.41 |

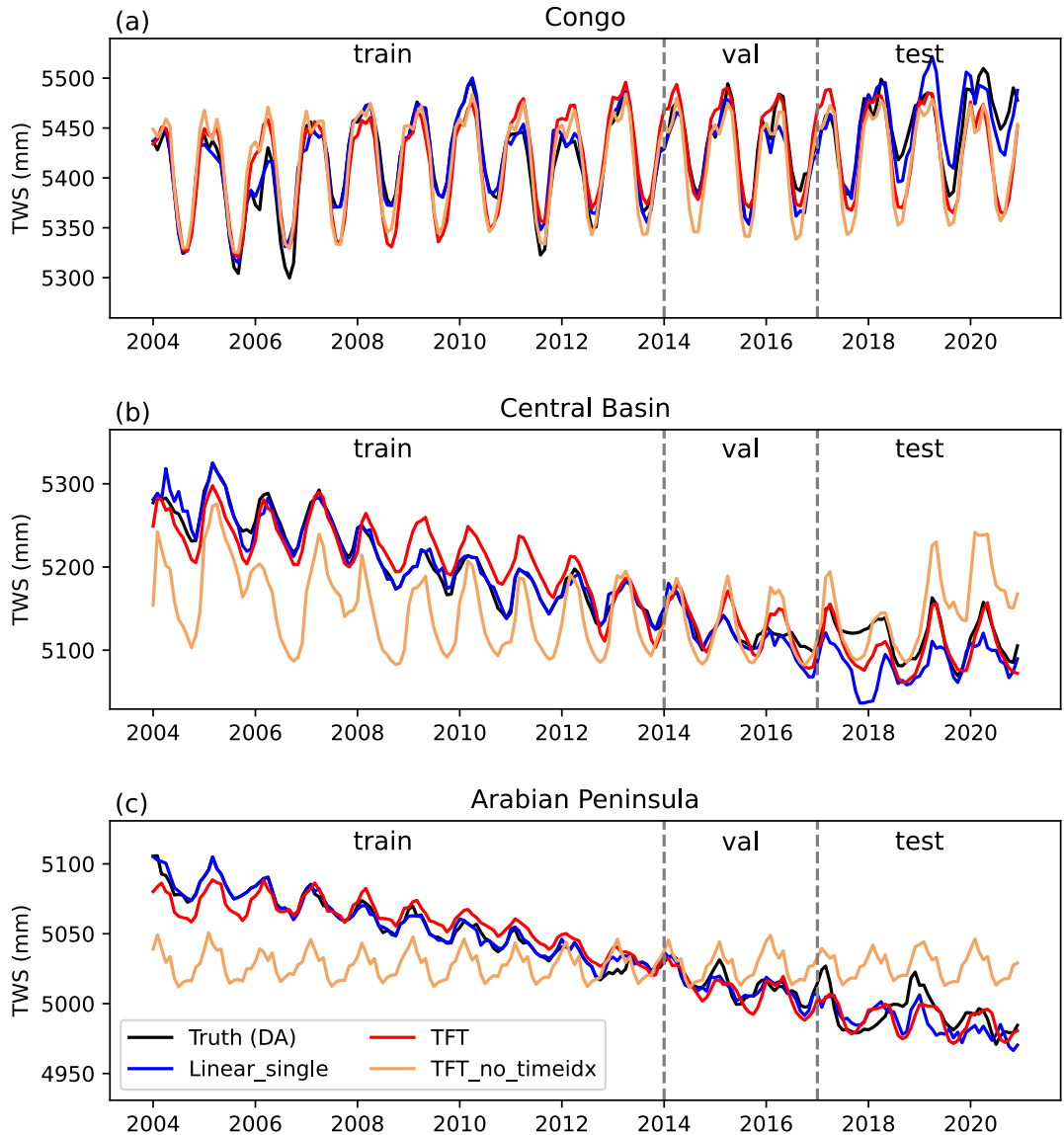

**Figure 7**. Examples of monthly time series of TWS for (**a**) Congo basin, (**b**) Central basin, and (**c**) Arabia Peninsula. Time series plots include TWS from DA dataset, Linear_single predicted TWS, along with the two TFT simulated TWS with and without global timestamp embedding.

### 3.6 Is training data size a limiting factor for LSTM and TFT?

According to scaling laws in deep learning, increasing the amount of data used for training can largely improve model performance, as large dataset provide better generalization and enhance the ability to model complex and nonstationary patterns. However, for time series tasks, expanding the training set –especially by increasing temporal resolution – introduces additional challenges. Higher-resolution data may contain more intricate temporal dependencies, increased noise, and short-term variations that complicate learning, potentially offsetting the benefits of larger training sizes.

We conduct an experiment where we switch from monthly time series to daily time series, inherently inducing finer scale patterns and greater variability. As a result, the training set for the globally trained LSTM and TFT models increases substantially – from 55620 to 375435 data points. Since predicting a single-day TWS target based on a full year of daily inputs may be overly sensitive to noise, we apply a 30-day moving average filter to the target variable. This procedure ensures that the model is predicting a smoothed 30-day averaged TWS rather than an individual, noisy daily value. As shown in Figure 8, for both the OL and DA datasets, neither LSTM nor TFT outperforms Linear_single even when trained with a correspondingly larger dataset. However, the performance of LSTM on the OL dataset appears to benefit from daily input, with the performance gap between LSTM and Linear_single narrows considerably, the correlation metric of which even becomes statistically insignificant. Overall, this experiment highlights that the size of training data is not the primary factor limiting LSTM and TFT relative to Linear_single benchmark.

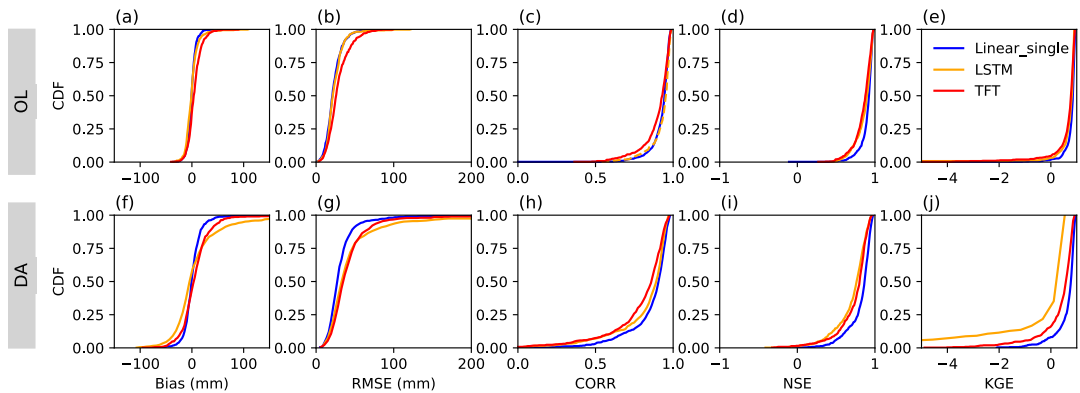

**Figure 8**. As in Figure 1 but for results trained on daily time-varying features with a 30-day moving average filter applied to the target TWS.

### 3.7 Do non-deep learning machine learning models provide an advantage over linear regression?

One might argue that while DL models struggle with global-scale heterogeneity across basins, such an observation does not automatically establish linear regression as the optimal benchmark, as other ML approaches exist that better handle basin-specific small datasets while capturing nonlinear feature-target relationships. In particular, tree-based models such as Random Forest (RF) and Light Gradient-Boosting Machine (LightGBM) are well suited for structured data, robust to outliers, and capable of capturing complex interactions between features (Grinsztajn et al., 2022; Ke et al., 2017), which could be advantageous given the potential nonlinearities in TWS dynamics.

To test the hypothesis that RF and LightGBM models may outperform Linear_single, we implement basin-specific RF and LightGBM models applied on monthly regression task with 12 months sequence length, optimizing hyperparameters for each basin independently (see Methods). Figure 9 compares the performance of these models against Linear_single across all basins. The results suggest that, besides the bias term – where RF does not significantly differ from Linear_single in terms of average performance – Linear_single consistently outperforms both tree-based models across all other evaluation metrics, particularly in correlation and NSE, for the OL dataset. For the DA dataset, LightBGM exhibits more comparable performance to Linear_single as it yields insignificant difference in bias, correlation, and KGE metrics. The performance gap between Linear_single and tree-based models is most pronounced in basins where the test period distribution differs significantly from the training period. This reinforces the observation that RF and LightGBM are particularly susceptible to distributional shifts, a challenge commonly observed in basins where substantial human intervention affects freshwater availability. Such nonstationarity exacerbates generalization errors in RF and LightBGM, even when a trend feature is incorporated. In this setting, where TWS dynamics exhibit a predominantly linear relationship with lagged precipitation, temperature, LAI, and SSMC, increasing model complexity doesn't necessarily enhance predictive skill, where the effectiveness of nonlinear approaches is constrained by the underlying structure of the feature-target relationship.

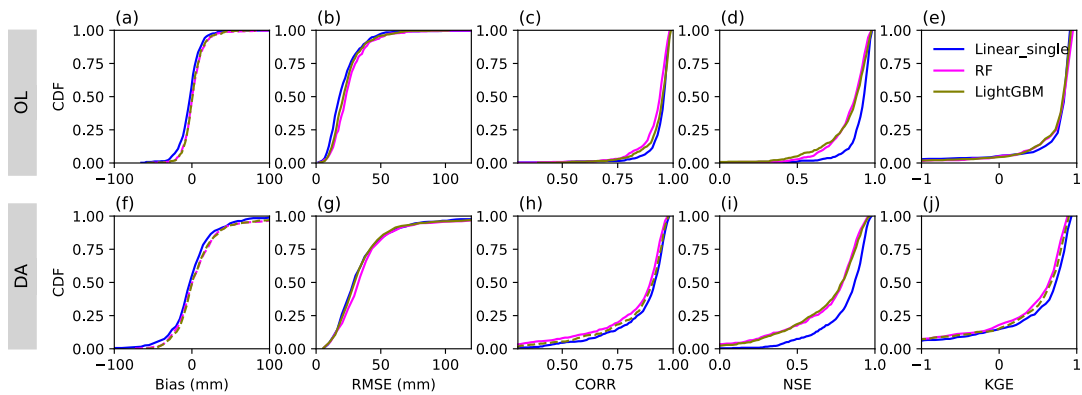

**Figure 9.** As in Figure 1 but for Random Forest (RF) and Light Gradient-Boosting Machine (LightGBM) compared to Linear_single.

## 4 Discussion and Conclusions

This study explores the effectiveness of DL models for hydrological time series prediction and highlights the importance of carefully selecting benchmark models. Our findings show that a simple linear model incorporating seasonal and trend components provides a strong baseline, performing competitively against widely used DL architectures such as LSTMs and Transformers specifically in the task of predicting TWS. However, we do not suggest that linear models are inherently superior across all hydrological applications. Instead, we emphasize the value of using simple, interpretable, and well-established benchmark models when developing or applying DL methods, ensuring that the performance gains from more complex models are well-justified. Although DL models can offer advantages in other hydrological applications, such as those involving large-scale spatial dependencies, the extraction of complex patterns from heterogeneous data sources, transfer learning scenarios where models trained on data-rich regions are applied to data-scarce regions, and remote sensing-based approaches that leverage foundation models to process massive volumes of satellite data (Fibaek et al., 2024; Jakubik et al., 2023; Spradlin et al., 2024; Szwarcman et al., 2024), our study highlights the challenges DL models can face in this particular task and datasets.

Our comparison of model performance across datasets with different levels of nonstationarity (OL vs. DA) reveals that all models, to varying degrees, struggle with learning from more nonstationary data. Although the linear model – despite its assumption of a simple linear trend – performs best on the DA dataset, and the TFT benefits from global timestamp embeddings to capture some aspects of nonstationarity, both models exhibit reduced skill compared to their performance on the OL dataset, which more closely resembles the natural hydrological system. The LSTM, in particular, struggles to capture the patterns in both OL and DA datasets. These findings suggest that model performance may be constrained by the available input features and that further methodological refinements or data preprocessing strategies could enhance the learning in a nonstationary environment. More broadly, this study highlights the need for benchmark datasets that capture both natural variability and human influences on hydrological fluxes. Such datasets are crucial for supporting the development and evaluation of models that better represent these effects on freshwater resources. Ultimately, improving data-driven hydrological modeling under changing conditions is crucial for applications such as identifying hydrological extremes, assessing water availability, and informing water resource management decisions.

## Acknowledgments

The research was supported by the NASA-sponsored NASA Water Insight project. Computational resources were provided by NASA's Center for Climate Simulation (NCCS).

## Open Research

HydroGlobe datasets and the code that are used for training and perform the data-driven models can be accessed from the Johns Hopkins Research Data Repository at the following URL: https://archive.data.jhu.edu/privateurl.xhtml?token=8c9e19b2-cf63-4e41-842e-73cd409be21f.